%% file: root.tex
\documentclass[letterpaper, 10 pt, conference]{ieeeconf}  

\IEEEoverridecommandlockouts                              

\input{packages.tex}

\input{Commands.tex}
\usepackage{hyperref}
\hypersetup{
    colorlinks=true, 
    linkcolor=black, 
    citecolor=green, 
    urlcolor=red,   
    breaklinks=true  
}
\usepackage[T1]{fontenc}
\usepackage{glossaries}
\glsresetall
\loadglsentries{acronyms}

\usepackage{algorithm,algpseudocode}

\usepackage[normalem]{ulem}

\title{\LARGE \bf
Federated Learning with Heterogeneous Data Handling for Robust Vehicular Object Detection
}

\author{%
\authorblockN{Ahmad Khalil\authorrefmark{1}\textsuperscript{\textbullet},
Tizian Dege\authorrefmark{1}\textsuperscript{\textbullet},
Pegah Golchin\authorrefmark{1},\\
Rostyslav Olshevskyi\authorrefmark{2},
Antonio Fernandez Anta\authorrefmark{3},
Tobias Meuser\authorrefmark{1}}
\authorblockA{\authorrefmark{1}Multimedia Communications Lab, Technical University of Darmstadt, Germany\\}
\authorblockA{\authorrefmark{2}Department of Electrical and Computer Engineering (ECE), Rice University, USA\\}
\authorblockA{\authorrefmark{3}IMDEA Networks Institute, Spain\\
Contact: ahmad.khalil@kom.tu-darmstadt.de
}
\thanks{\textsuperscript{\textbullet}Authors contributed equally to the paper.}
}

\begin{document}
\maketitle
\begin{abstract}
  \input{Sections/0-Abstract}

\end{abstract}
\input{Sections/1-Introduction}
\input{Sections/2-Heterogeneity_Handling}
\input{Sections/3-Evaluation}

\input{Sections/4-Conclusion}
\input{Sections/5-Achnowledgement}
\thispagestyle{empty}
\pagestyle{empty}
\bibliographystyle{unsrt}

\end{document}

%% file: packages.tex
\usepackage[utf8]{inputenc}
\usepackage{mathtools}
\usepackage{comment}
\usepackage{multicol, blindtext}
\usepackage{booktabs}
\usepackage{multirow}
\usepackage{url}
\usepackage{listings}
\usepackage{accents}
\usepackage{cite}
\usepackage{amsmath,amssymb,amsfonts}
\usepackage{graphicx}
\usepackage{textcomp}
\usepackage{xcolor}
\usepackage{algorithm}
\usepackage[noend]{algpseudocode}
\usepackage{subcaption}
\usepackage{amsmath}
\usepackage{tabularx}

\algdef{SE}[SUBALG]{Indent}{EndIndent}{}{\algorithmicend\ }%
\algtext*{Indent}
\algtext*{EndIndent}

%% file: Commands.tex
\usepackage{xcolor}  
\usepackage{todonotes}
\usepackage{ifthen}

\newcommand{\eg}{e.g.,\xspace}
\newcommand{\ie}{i.e.,\xspace}

\AtBeginDocument{%
  \providecommand\BibTeX{{%
    \normalfont B\kern-0.5em{\scshape i\kern-0.25em b}\kern-0.8em\TeX}}}

\newcommand{\todoinline}[2][]{
  \ifthenelse{\equal{#1}{}}{
    \todo[inline]{\textbf{TODO: #2}}
  }
  {
    \todo[inline]{\textbf{TODO (#1): #2}}
  }
}

%% file: acronyms.tex
\newacronym{FL}
{FL}{Federated Learning}

\newacronym{ADAS}
{ADAS}{Advanced Driver Assistance System}

\newacronym{IoU}
{IoU}{Intersection Over Union}

\newacronym{LiDAR}
{LiDAR}{Light Detection and Ranging}

\newacronym{ML}
{ML}{Machine Learning}

\newacronym{RSU}
{RSU}{Road Side Unit}

\newacronym{IID}
{IID}{Independent and Identically Distributed}

\newacronym{non-IID}
{non-IID}{Non-Independent and Identically Distributed}

\newacronym{FedAvg}
{FedAvg}{Federated Averaging}

\newacronym{Fed2A}
{Fed2A}{Federated Learning Mechanism in Asynchronous and Adaptive Modes}

\newacronym{FedBiKD}
{FedBiKD}{Federated Bidirectional Knowledge Distillation}

\newacronym{mAP}
{mAP}{mean Average Precision}

\newacronym{CP}
{CP}{Collective Perception}

\newacronym{V2X}
{V2X}{Vehicle-to-everything}

\newacronym{DNN}
{DNN}{Deep Neural Network}

%% file: Sections/0-Abstract.tex
\label{abs}
In the pursuit of refining precise perception models for fully autonomous driving, continual online model training becomes essential.
\gls{FL} within vehicular networks offers an efficient mechanism for model training while preserving raw sensory data integrity.
Yet, \gls{FL} struggles with non-identically distributed data (\eg quantity skew), leading to suboptimal convergence rates during model training.
In previous work, we introduced FedLA, an innovative Label-Aware aggregation method addressing data heterogeneity in \gls{FL} for generic scenarios.

In this paper, we introduce FedProx+LA, a novel \gls{FL} method building upon the state-of-the-art FedProx and FedLA to tackle data heterogeneity, which is specifically tailored for vehicular networks.
We evaluate the efficacy of FedProx+LA in continuous online object detection model training.
Through a comparative analysis against conventional and state-of-the-art methods, our findings reveal the superior convergence rate of FedProx+LA.
Notably, if the label distribution is very heterogeneous, our FedProx+LA approach shows substantial improvements in detection performance compared to baseline methods, also outperforming our previous FedLA approach.
Moreover, both FedLA and FedProx+LA increase convergence speed by 30\% compared to baseline methods.

%% file: Sections/1-Introduction.tex
\section{Introduction}
\label{sec:intro}
\glsresetall
To enable fully autonomous driving, \glspl{ADAS} must exhibit solid reliability across a spectrum of scenarios~\cite{van2018autonomous}. 
Conventionally, sensory data collected from the vehicle's sensors, including \glspl{LiDAR}, Cameras, and Radars, is utilized by perception models, such as \glspl{DNN}, to construct a thorough representation of the surroundings -- a local environment model~\cite{liu2020computing}. 
The environmental model's accuracy significantly influences vehicle control decisions.
Yet, solely training models centrally with constrained data compromises reliability~\cite{schiegg2020collective}.
To overcome these limitations and enhance adaptable perception models, the crucial transition to continuous online model training becomes apparent~\cite{khalil2022situational}.
This shift is pivotal for establishing a robust foundation for fully autonomous driving.
FL has recently emerged as a technique for training models without requiring clients to exchange data with a central entity during the training process~\cite{mcmahan2017communication, mammen2021federated, khalil2022situational}.
\begin{figure}[t]
\centering\includegraphics[width=0.3\textwidth]{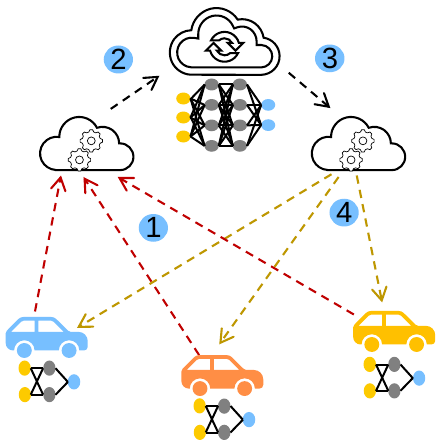}
\caption{Online training of the general perception model using FL \cite{khalil2022situational}. Each communication round involves (1) transmission of local updates to the edge server (\eg \gls{RSU}), where intermediate aggregation occurs. (2) The edge servers transfer the aggregated model to the central aggregation server for final aggregation. (3) and (4) The aggregated model is subsequently transferred back to individual clients (vehicles).}
  \label{fig:FLForTraining}
  \vspace{-1.5em}
\end{figure}
\gls{FL} enables training on distributed client data at each communication round.
As illustrated in Figure \ref{fig:FLForTraining}, client updates are aggregated iteratively to converge towards the global optimum.
Notably, these updates are much smaller than clients' entire datasets, a departure from conventional training methods requiring extensive data exchange.
\gls{FL}'s ability to support continuous online training and reduce data exchange makes it ideal for vehicular \gls{ML} applications.
Its adoption in vehicular networks has sparked interest, where vehicles serve as clients, generating perception data without overwhelming the communication network.
Additionally, \glspl{RSU} or cloud servers can coordinate training by initiating and aggregating model updates from vehicles during each communication round~\cite{pervej2023resource}.
\gls{FL} performs well with \gls{IID} data but struggles with \gls{non-IID} data, where data heterogeneity among clients poses challenges to its performance \cite{vahidian2023rethinking, ye2023heterogeneous}.
In our previous work~\cite{khalil2023label}, we explored the impact of \gls{non-IID} data on training processes, categorizing strategies into \textbf{Data-based} and \textbf{Parameter-based} approaches.
FedLA, one of our previous contributions \cite{khalil2023label}, addressed label distribution skew using a data-based approach.
In that work, we evaluated FedLA using common \gls{FL} datasets, demonstrating its effectiveness in mitigating non-IIDness in non-vehicular scenarios.

In this paper, we focus on \gls{FL} in vehicular scenarios, \eg propose a novel \gls{FL} method called FedProx+LA, and evaluate the applicability of our previous work in vehicular scenarios.
Our FedProx+LA approach combines our previous work with the capabilities of FedProx~\cite{li2020federated} to address the data heterogeneity issue explicitly in vehicular scenarios.
FedProx+LA represents a hybrid approach that takes into account both data characteristics and model parameters.
Furthermore, we evaluate the effectiveness of FedLA and FedProx+LA in the context of online model training for object detection.
To assess their performance, we conduct a comparative analysis against \gls{FedAvg}~\cite{mcmahan2017communication} as a traditional aggregation method, our newly introduced method FedAvgL, and FedProx~\cite{li2020federated} as a state-of-the-art alternative.
We evaluate the approaches using the famous  realistic vehicular perception dataset (Nuscenes).
While our analysis demonstrates considerable improvements in detection performance and convergence rates already with FedLA compared to FedAvg, FedAvgL, and FedProx, the detection performance was further improved by our FedProx+LA approach tailored for perception in vehicular networks.
This improvement is underscored by a substantial increase in the detection performance measured by \gls{mAP}, with FedLA and FedProx+LA achieving enhancements of 5\% and 6\%, respectively.
Additionally, FedLA and FedProx+LA exhibit remarkable improvements in convergence speed, achieving a significant 30\% enhancement compared to the baseline methods.
The comprehensive source code, covering various data distribution scenarios and corresponding experimental results across different performance metrics, is openly accessible \footnote{\url{https://github.com/TixXx1337/Federated-Learning-with-Heterogeneous-Data-Handling-for-Robust-Vehicular-Perception}}.

The subsequent sections of this paper are structured as follows: Section \ref{sec:rw} comprehensively reviews recent literature on addressing heterogeneity in the model training process within vehicular networks using \gls{FL}.
In Section \ref{sec:fedla}, we introduce FedProx+LA.
In Section \ref{sec:eval}, we conduct a comprehensive empirical assessment of FedLA and FedProx+LA, exploring their performance across diverse experimental configurations. Finally, we conclude our work with Section \ref{sec:conc}.

%% file: Sections/2-Heterogeneity_Handling.tex
\section{Related Work}
\label{sec:rw}
Heterogeneous data distribution among clients presents a significant challenge in \gls{FL}, potentially degrading its performance \cite{kairouz2021advances, li2022federated, ye2023heterogeneous, vahidian2023rethinking}.
Initial solutions, like \gls{FedAvg}~\cite{mcmahan2017communication}, addressed data size differences but did not consider other \gls{non-IID} scenarios, neglecting scenarios with data heterogeneity in \gls{FL}.
Differences in local objectives and long local epochs may lead to the divergence of the local models from the global model, hindering convergence or causing divergence.

In Zhao et al.'s study~\cite{zhao2018federated}, they observed a notable decrease in accuracy when models were trained on highly skewed non-IID data.
To mitigate this, they proposed a strategy of creating a small subset of globally shared data among edge devices.
Li et al.~\cite{li2019convergence} explored the impact of \gls{non-IID} settings on model performance in \gls{FedAvg}, highlighting how data heterogeneity significantly slowed convergence.
These findings underscore the necessity for tailored approaches to address \gls{non-IID} data challenges in \gls{FL} scenarios.

FedAvgM, proposed by Hsu et al.~\cite{hsu2019measuring}, aimed to enhance model aggregation in \gls{FL} by incorporating momentum during the global model update on the server.
This momentum-based approach aimed to improve aggregation efficiency and effectiveness.
Conversely, FedNova, introduced by Wang et al.~\cite{wang2020tackling}, normalized local updates before aggregation to stabilize the process, addressing issues related to varying scales and distributions of updates.
FedProx~\cite{li2020federated} addressed variable local updates in statistically heterogeneous settings by adding a proximal term to the local problem, expediting convergence in such scenarios.
MOON~\cite{li2021model} introduced model-contrastive \gls{FL}, aligning local and global model representations to enhance \gls{FL} system performance.

In the realm of vehicular applications, the challenge of heterogeneity in \gls{FL} has prompted innovative solutions.
Liu et al.~\cite{liu2022fed2a} addressed the direct impact of local parameter heterogeneity by introducing \gls{Fed2A}.
This approach aimed to mitigate the effects of heterogeneity to enhance the model performance by leveraging the temporal and informative attributes of local parameters.
The proposed aggregation scheme incorporated considerations for the \textit{freshness} of received local models through the application of temporal weights.
Additionally, it emphasized \textit{consistency} by penalizing stale models, contributing to a more robust and adaptive \gls{FL} framework tailored to vehicular applications.
Shang et al.~\cite{shang2023fedbikd} presented the \gls{FedBiKD} framework as a targeted solution to address the issue of local deviation in vehicular scenarios.
In \gls{FedBiKD}, global knowledge played a pivotal role in guiding local training, while common knowledge derived from an ensemble of local models was employed to fine-tune the aggregated global model.

\section{Federated Label-Aware Averaging in Vehicular Networks}
\label{sec:fedla}
In this section, we present our FedProx+LA approach, which combines our previous FedLA approach for generic \gls{FL} with the FedProx~\cite{li2020federated} approach from related work, which was specifically tailored for vehicular scenarios.
In the following, we first give a short introduction to FedLA and then describe our new FedProx+LA approach.

\subsection{FedLA} 
In addressing label distribution skew, our prior research, outlined in~\cite{khalil2023label}, introduced FedLA.
The core of FedLA, as depicted in Algorithm \ref{algo:FedLA}, illustrates the server's computation to determine client weights based on their accumulated weights across all labels.
\begin{algorithm}[t]
\caption{Weight calculation for FedLA in every communication round. Here, $k$ represents the number of participating clients, $n_l$ denotes the total number of labels (classes), and $S(c_i)$ is the number of samples held by client $i$ in the communication round. For a specific label with index $j$ (i.e., $l_j$), $S(l_j)$ denotes the number of samples labeled $l_j$, and $S(c_i, l_j)$ is the number of samples with label $l_j$ at client $c_i$~\cite{khalil2023label}.}
\label{algo:FedLA}
\begin{algorithmic}[1]
\State Initialize: $k$, $S(c_i, l_j)$ (for all clients participating in the communication round) 
    \For {each label $l_j$}
        \State $S(l_j) = \sum_{i=1}^{k} S(c_i,l_j)$ \Comment{calculate total number of samples of that specific label}
        \For {each client $c_i$}
            \State $W(c_i,l_j) = \dfrac{S(c_i,l_j)}{S(l_j)}$ \Comment{compute client weight per label}
        \EndFor
    \EndFor
    
    \For {each client $c_i$}
        \State $W(c_i) = \sum_{j=1}^{n_l} W(c_i,l_j)$ \Comment{compute client weight w.r.t. all labels}
        \State $W_{FedLA}(c_i) = \dfrac{W(c_i)}{\sum_{x=1}^{k} W(c_x)}$ \Comment{compute client update weight}
    \EndFor
\end{algorithmic}
\end{algorithm}
In our investigation of FedLA, we utilized two commonly used datasets in FL: Emnist-balanced~\cite{cohen2017emnist} and CIFAR-100~\cite{cifar100}.
It's important to recognize that FedLA's effectiveness relies on the assumption that the central server has comprehensive insights into label distributions throughout every round of communication, encompassing all participating clients.
However, it's worth acknowledging that this assumption may not always be practical in certain \gls{FL} applications, \eg due to privacy considerations.
Nonetheless, in \gls{FL} applications like training object detection models for vehicular contexts, where clients share lightweight anonymized label information, such as car presence or vegetation, along with their frequencies, the privacy risks may not be significant.
This observation motivates our exploration of FedLA's performance in the context of online training for vehicular object detection models.

\subsection{FedProx+LA} 
\begin{figure}[t]
\centering
    \includegraphics[width=0.5\textwidth]{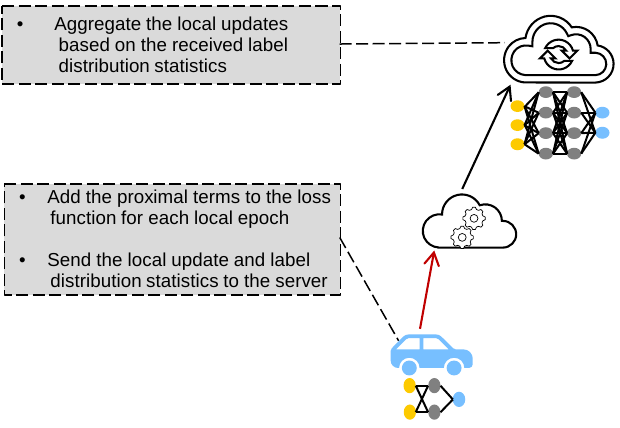}
\caption{FedProx+LA method: At each global epoch (communication round), every client (vehicle) incorporates the proximal term $\mathcal{L} = \mathcal{L}_{ce} + \mu\mathcal{L}_{prox}$ into its local update and transmits it alongside the label distribution statistics. Upon receiving the model updates along with the label distribution statistics, the aggregation server considers them to generate the aggregated global model.}
  \label{fig:FedProx+LA}
  \vspace{-1.5em}
\end{figure}
FedProx~\cite{li2020federated} emerges as a leading-edge \gls{FL} technique, tailored to tackle data heterogeneity.
It introduces a proximal term into local updates and leverages client sample sizes similar to \gls{FedAvg}. 
In this study, we introduce FedProx+LA, a novel extension of FedProx.
Illustrated in Figure \ref{fig:FedProx+LA}, FedProx+LA not only integrates proximal terms into client-side operations but also leverages label distribution knowledge (similar to FedLA) during model aggregation on the central server.
This integration promises enhanced performance and robust adaptability to data heterogeneity.
We evaluate the FedProx+LA method in Section \ref{sec:eval}, where it demonstrates exceptional performance, underscoring its efficacy in practical \gls{non-IID} settings.

%% file: Sections/3-Evaluation.tex
\section{Evaluation}
\label{sec:eval}
\begin{figure*}
  \centering
  \begin{subfigure}[b]{0.45\textwidth}
    \centering
    \includegraphics[width=\textwidth]{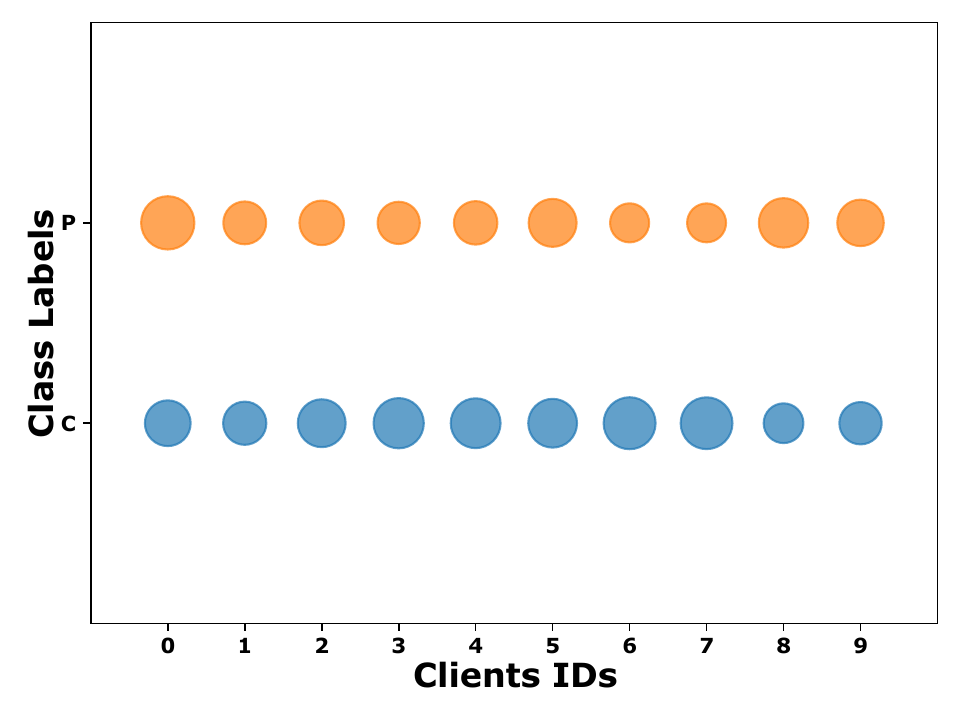}
    \caption{}
    \label{fig:DataDistributionIID}
  \end{subfigure}%
  \begin{subfigure}[b]{0.45\textwidth}
    \centering
    \includegraphics[width=\textwidth]{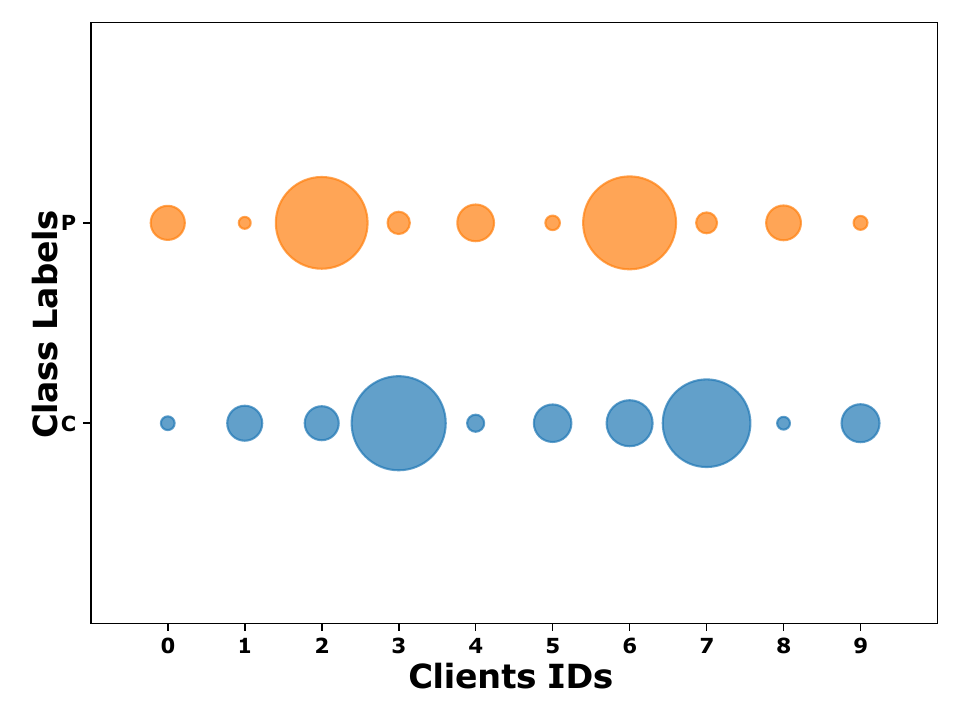}
    \caption{}
    \label{fig:DataDistributionNonIID}
  \end{subfigure}
  \caption{Different data distribution schemes are illustrated with Pedestrian (P) and Car (C) class labels. (a) depicts \gls{IID} data distribution across clients. (b) showcases \gls{non-IID} data distribution using the Dirichlet distribution. Both distributions ensure equal sample sizes across clients.}
  \label{fig:DataDistribution}
    \vspace{-1.5em}
\end{figure*}
In this section, we perform thorough experiments on a widely utilized public dataset to evaluate the efficacy of FedLA and FedProx+LA methods.
Additionally, we incorporate comparisons with alternative baselines to illustrate the effectiveness of our approaches.
\subsection{Experimental Setups}
\subsubsection{Data Set and Data Distribution}
The assessment focuses on online training of a global object detection model using vehicles (e.g., cars) as \gls{FL} clients.
The NuScenes Dataset~\cite{Caesar.2019} is chosen for its widespread use in vehicular perception research.
The study involves a total of $N=10$ clients (vehicles), each assigned 10 scenes.
Subsequently, various preprocessing procedures are implemented, including the exclusion of images lacking labels.
Additionally, specific steps are undertaken corresponding to each data distribution condition, ensuring equal sample sizes across the clients. 
To ensure \gls{IID} conditions in the NuScenes dataset, we randomly distribute scenes as complete blocks, ensuring both Pedestrian (P) and Car (C) classes are available for each image (Figure~\ref{fig:DataDistributionIID}). Conversely, for data heterogeneity or \gls{non-IID} scenarios, we use the Dirichlet distribution~\cite{elflein2023out}. Unlike common uses of the Dirichlet distribution, realistic object detection tasks exhibit correlations between all labels. We assign a preference for one class to each client and distribute scenes accordingly, maintaining identical sample counts and diverse label counts (Figure~\ref{fig:DataDistributionNonIID}), simulating real-world scenarios.
\subsubsection{Implementation Details}
At each global epoch, 50\% of clients ($C=0.5$) are randomly selected, with a configuration of $E_g=50$ global epochs (communication rounds).
During each global epoch, selected clients undergo local training for $E_l=10$ local epochs.
The chosen model is a custom Yolov8n~\cite{jocheryOLObyultralytics2023} with 3011238 parameters, size $\sim$6.2 MB, and modified for two classes (pedestrian, car).
Default learning rates ($lr_0=0.01, lr_f=0.01$) and the ADAM optimizer are used.
Validation employs K-fold cross-validation ($K=5$), with each data chunk representing 20\% of the dataset. 
Four chunks are designated for training, and one for testing, with reported results being an average across the 5 splits.
\subsubsection{Baselines}
To study the efficacy of our proposed methods, we conduct a comparative evaluation involving FedLA, and FedProx+LA against both a centralized training paradigm and two prominent \gls{FL} algorithms. 
\begin{itemize}
    \item \textit{Centralized Training (Central):} The object detection model is trained on the complete dataset aggregated from all clients on a central server.
    Training spans a cumulative epoch count equal to the total epochs in the \gls{FL} setups, specifically $E=50 \times 10 = 500$.
    \item \textit{FedAvg:} FedAvg~\cite{mcmahan2017communication} represents a widely recognized \gls{FL} method that leverages model averaging based on individual client sample sizes.

    \item \textit{FedProx:} FedProx~\cite{li2020federated} introduces a proximal term to local updates for enhanced performance (\ie $\mathcal{L}$ = $\mathcal{L}_{ce}$ + $\mu\mathcal{L}_{prox}$).
    In our experiments, we set the parameter $\mu$ to 0.01.

    \item \textit{FedAvgL:} FedAvgL is a refined variant of FedLA designed to disclose a reduced volume of statistical information. In FedAvgL, participating clients share the total number of labels present in all the samples used for training their local models, irrespective of the specific label indices (classes). Based on this, the weights are calculated, which mitigates the adverse impact of samples with either empty or extremely low label counts.
\end{itemize}


\subsubsection{Evaluation Metrics}
\begin{itemize}
    \item \textit{mAP (Mean Average Precision)}: The \gls{mAP} stands as a widely employed metric to assess the detection performance of a trained model \cite{mahaur2023small}.
    It comprehensively accounts for both Precision (P) and Recall (R) through the following formulations:
    \begin{align*}
    P= \frac{TP}{TP + FP}&&&&&&R= \frac{TP}{TP + FN}
    \end{align*}
    Here, TP denotes the number of detections present in both the ground truth and the detection results.
    FP represents the number of detections in the detection results only, not found in the ground truth.
    FN denotes the number of detections in the ground truth only, absent in the detection results.

    The Average Precision (AP) is computed as the average precision at different recall values obtained from the Precision-Recall (PR) curve under a given threshold:

    \[AP = \sum_{k}(R_{k+1} - R_k) \cdot \max_{\hat{R} \geq R_{k+1}}P(\hat{R})\]

    where \(P(\hat{R})\) represents the measured Precision at \(\hat{R}\).
    The \gls{IoU} quantifies the degree of overlap between the predicted box and the actual box~\cite{rezatofighi2019generalized}, with a predefined threshold in our study set to \gls{IoU} \(\geq\) 0.5.

    Finally, the \gls{mAP} is determined as the mean of AP over all classes:

    \[mAP = \frac{1}{n} \sum_{i=1}^{n}AP_i\]

    where \(n\) is the number of classes, and \(AP_i\) signifies the AP at class \(i\).
    
    \item \textit{$E_g$@mAP}: To measure the communication efficiency and the convergence rate \cite{shang2023fedbikd}. 
    It refers to the number of global epochs required to achieve the desired \gls{mAP} for the first time.
\end{itemize}

\subsection{Results and Discussion}

\begin{figure}[t]
\centering
    \includegraphics[width=0.45\textwidth]{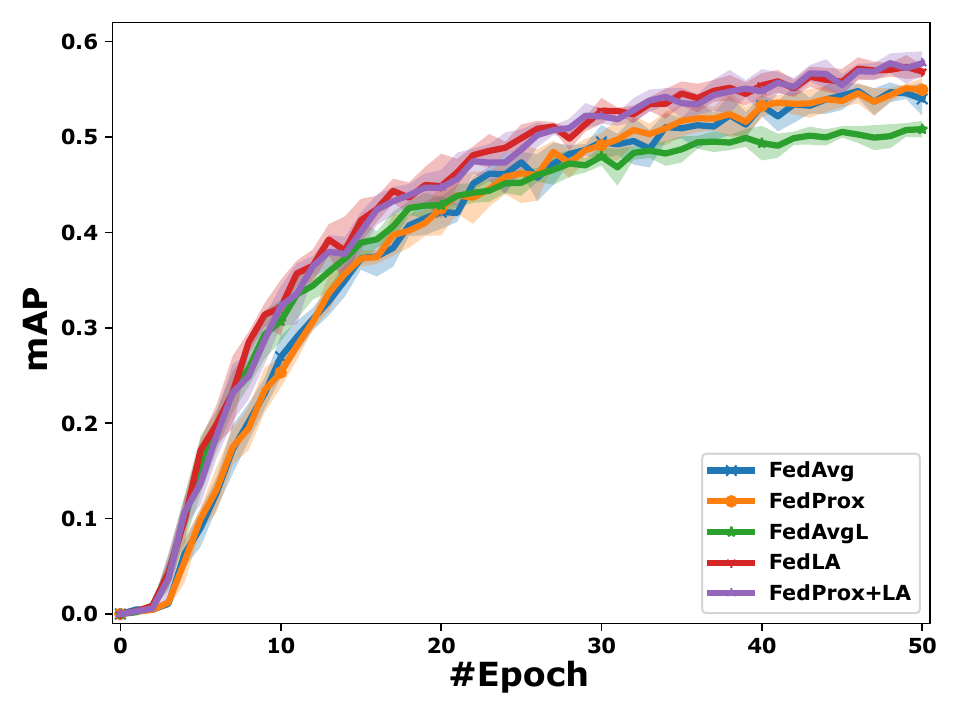}
  \caption{Comparison of \gls{mAP} (\gls{IoU} \(\geq\) 0.5) across various methods in the \gls{non-IID} setting.} 
  \label{fig:map50nonIID}
\end{figure}

\begin{figure}[t]
\centering
\includegraphics[width=0.425\textwidth]{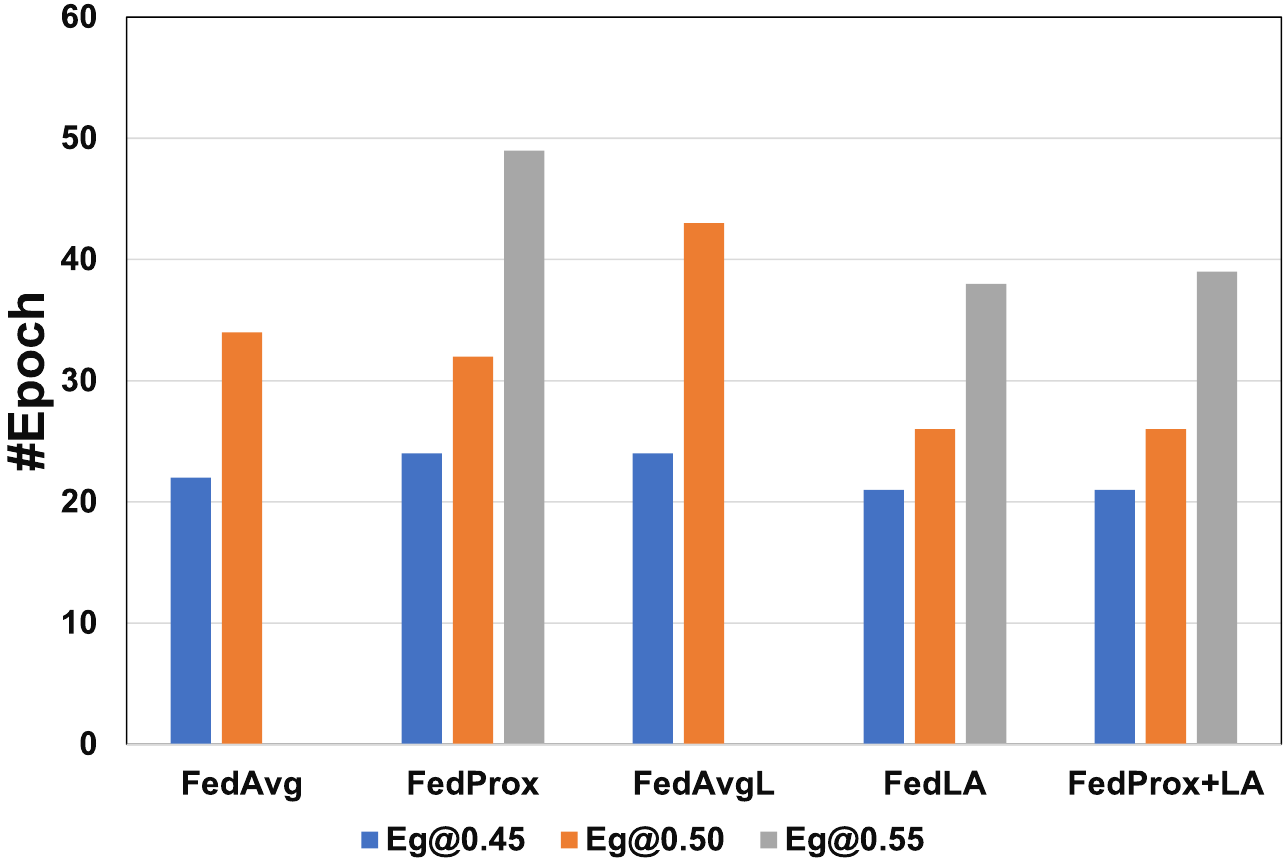}
  \caption{Comparison of convergence rates in the \gls{non-IID} setting: FedAvg and FedAvgL fail to reach the target \gls{mAP} of 0.55 within global epochs $E_g = 50$. While FedProx+LA and FedLA achieve a considerable convergence rate.} 
  \label{fig:converganceRatesnonIID}
      \vspace{-1.5em}
\end{figure}

\begin{table*}
    \centering
    \caption{Comparison of \gls{mAP} (\gls{IoU} \(\geq\) 0.5) and the number of global epochs required to achieve various target \gls{mAP} values across different methods. The value in parentheses () indicates the speedup relative to \gls{FedAvg}. "x" denotes that the target \gls{mAP} was never attained. If "x" appears for \gls{FedAvg}, then the speedup is calculated relative to FedProx as a baseline. }
    \label{tab:performance_metrics}
    \begin{tabular}{lccccccc}
        \toprule
        \textbf{Data Distribution} & & \textbf{Central} & \textbf{FedAvg} & \textbf{FedProx} & \textbf{FedAvgL} & \textbf{FedLA} & \textbf{FedProx+LA} \\
        \midrule
        \multirow{4}{*}{IID} & mAP & 0.779 & 0.597 & 0.590 & \textbf{0.597} & \textbf{0.597} & 0.595 \\
                             & $E_g$@0.45 & - & 14 (1.0$\times$) & 15 (0.93$\times$) & 14 (1.0$\times$) & 14 (1.0$\times$) & 14 (1.0$\times$)\\
                             & $E_g$@0.50 & - & 20 (1.0$\times$) & 21 (0.95$\times$) & \textbf{19 (1.05$\times$)}\ & \textbf{19 (1.05$\times$)} & 20 (1.0$\times$)\\
                             & $E_g$@0.55 & - & 30 (1.0$\times$) & 29 (1.03$\times$) & \textbf{28 (1.07$\times$)} & 29 (1.03$\times$) & 30 (1.0$\times$)\\
        \midrule
        \multirow{4}{*}{non-IID} & mAP & 0.779 & 0.54 & 0.549 & 0.508 & 0.569 & \textbf{0.577} \\
                             & $E_g$@0.45 & - & 22 (1.0$\times$) & 24 (0.92$\times$) & 24 (0.92$\times$) & \textbf{21 (1.05$\times$)} & \textbf{21 (1.05$\times$)}\\
                             & $E_g$@0.50 & - & 34 (1.0$\times$) & 32 (1.06$\times$) & 43 (0.79$\times$) & \textbf{26 (1.3$\times$)} & \textbf{26 (1.3$\times$)}\\
                             & $E_g$@0.55 & - & x & 49 (1.0$\times$) & x & \textbf{38 (1.29$\times$)} & 39 (1.26$\times$)\\
        \bottomrule
    \end{tabular}
\end{table*}
\subsubsection{Object Detection Performance}
The analysis of the object detection performance reveals several key insights.
Figure \ref{fig:map50nonIID} and Table \ref{tab:performance_metrics} provide comprehensive comparisons across different methods.
In the \gls{IID} setting, all methods exhibit similar performance characteristics, showing comparable \gls{mAP} values.
Across all of the compared approaches, there is a notable decrease in \gls{mAP} scores when switching from the \gls{IID} scenario to the \gls{non-IID} scenario.
This is expected, given that, the data distributed among the clients in \gls{IID} setting exhibit minimal variations in label distribution (Figure \ref{fig:DataDistributionIID}).
However, in the \gls{non-IID} setting, our label-aware approaches (FedLA and FedProx+LA) demonstrate a clear advantage over other methods across all global training epochs.
FedLA achieves up to a $5\%$ improvement in \gls{mAP} compared to the baseline, while FedProx+LA achieves even higher enhancements, with improvements of up to $6\%$.
Interestingly, FedAvgL initially performs reasonably well, but its performance deteriorates notably after $E_g=25$, ultimately becoming the least effective method in the \gls{non-IID} setting, even underperforming FedAvg.
This arises from the emphasis solely on label counts, which may reduce the weight of client models with fewer total labels, even in cases where they hold marginal differences in label counts across different classes.
This influences the aggregated model to prioritize one class in a given global epoch and subsequently transition to the other class in the following epoch posing challenges to the convergence.
Similar to FedLA, FedProx+LA addresses this issue by taking into account the label count per class of selected clients, thereby balancing the weights and boosting the detection performance.
However, FedProx+LA additionally incorporates proximal terms to prevent deviation from the global optimum, resulting in a slight improvement in detection performance compared to FedLA.
Finally, both FedAvg and FedAvgL exhibit limitations in reaching higher target \gls{mAP} values in the \gls{non-IID} setting, indicating challenges in adapting to the \gls{non-IID} data distribution.

\subsubsection{Communication Efficiency and Convergence Rate}
The analysis of communication efficiency and convergence rate provides insight into the behavior of different methods across various settings, as depicted in Figure \ref{fig:converganceRatesnonIID} and Table \ref{tab:performance_metrics}.
In the \gls{IID} setting, all methods demonstrate relatively similar convergence patterns, although FedProx exhibits slightly slower convergence compared to other methods.
Across both \gls{IID} and \gls{non-IID} settings, fewer global training epochs are required to attain the target \gls{mAP} in the \gls{IID} scenario compared to the \gls{non-IID} scenario, indicating the impact of data distribution on convergence efficiency.
In the \gls{non-IID} settings, FedAvgL exhibits a notable deceleration in the convergence speed that impedes the trained model from reaching higher target \gls{mAP} values.
This is associated with the factors contributing to the previously discussed poor detection performance of this method.
On the other hand, FedLA demonstrates marginally greater efficiency compared to FedProx+LA in achieving the target \gls{mAP}.
However, the difference in efficiency between the two methods is relatively minor.
Moreover, as the target \gls{mAP} value increases in the \gls{non-IID} setting, both FedLA and FedProx+LA demonstrate substantial improvements in convergence speed, achieving a remarkable $30\%$ enhancement compared to the baseline methods.
These findings highlight the subtle relationship between communication efficiency, convergence rate, and data distribution characteristics, highlighting the effectiveness of data heterogeneity handling mechanisms in enhancing convergence rate, particularly in \gls{non-IID} scenarios.

%% file: Sections/4-Conclusion.tex
\section{Conclusion}
\label{sec:conc}
In our study, we highlighted the vital role of continual training for perception models in autonomous driving, exploring the efficacy of \gls{FL} within vehicular networks.
A key challenge we addressed was data heterogeneity, which can hinder the \gls{FL} process due to data disparities.
In this study, we proposed FedProx+LA, a hybrid method built on FedProx and FedLA, which considers both data characteristics and model parameters.
Conducting a comparative analysis, we evaluated FedProx+LA against methods like FedAvg, FedAvgL, FedProx, and our previous work FedLA.
Our results, particularly with the Nuscenes dataset, highlight the superior performance of FedProx+LA.
We observed significant enhancements in detection performance (6\% for FedProx+LA, 5\% for FedLA) and a remarkable 30\% increase in convergence speed over baseline methods.

%% file: Sections/5-Achnowledgement.tex
\section*{Acknowledgement}
\label{sec:ack}
This work has been funded by the German Research Foundation (DFG) within the Collaborative Research Center (CRC) 1053 MAKI.
The authors gratefully acknowledge the computing time provided to them on the high-performance computer Lichtenberg at the NHR Centers NHR4CES at TU Darmstadt.